\def\year{2022}\relax
\documentclass[letterpaper]{article} 
\usepackage{aaai22}  
\usepackage{times}  
\usepackage{helvet}  
\usepackage{courier}  
\usepackage[hyphens]{url}  
\usepackage{graphicx} 
\usepackage{natbib}  
\usepackage{caption} 
\DeclareCaptionStyle{ruled}{labelfont=normalfont,labelsep=colon,strut=off} 
\usepackage{algorithm}
\usepackage{algorithmic}
\usepackage{newfloat}
\usepackage{listings}
\floatstyle{ruled}
\newfloat{listing}{tb}{lst}{}
\floatname{listing}{Listing}
\title{Inferring Lexicographically-Ordered Rewards from Preferences}
\author{
    Alihan H\"uy\"uk\textsuperscript{\rm 1},
    William R. Zame\textsuperscript{\rm 2},
    Mihaela van der Schaar\textsuperscript{\rm 1\,2\,3}
}
\affiliations{
    \textsuperscript{\rm 1}University of Cambridge \\
    \textsuperscript{\rm 2}University of California, Los Angeles \\
    \textsuperscript{\rm 3}The Alan Turing Institute \\
    ah2075@cam.ac.uk,
    zame@econ.ucla.edu,
    mv472@cam.ac.uk

}

\usepackage{amssymb}
\usepackage{amsthm}
\usepackage{bm}
\usepackage{booktabs}
\usepackage{caption}
\usepackage{enumitem}
\usepackage{graphicx}
\usepackage{makecell}
\usepackage{mathtools}
\usepackage{microtype}
\usepackage{multirow}
\usepackage{subcaption}
\usepackage{tabularx}
\usepackage{url}
\usepackage{wrapfig}
\usepackage{xcolor}

\DeclareMathOperator{\argmax}{argmax}
\DeclareMathOperator{\softmin}{softmin}
\newcommand{\Ex}{\mathbb{E}}
\renewcommand{\Pr}{\mathbb{P}}

\allowdisplaybreaks
\ifodd 0 \newcommand{\com}[1]{{\color{red}#1}} \else \newcommand{\com}[1]{#1} \fi
\ifodd 0 \newcommand{\rev}[1]{{\color{blue}#1}} \else \newcommand{\rev}[1]{#1} \fi

\begin{document}
\maketitle

\begin{abstract}
	{Modeling the preferences of agents over a set of alternatives is a principal concern in many areas. The dominant approach has been to find a single reward/utility function with the property that alternatives yielding higher rewards are preferred over alternatives yielding lower rewards. However, in many settings, preferences are based on multiple---often competing---objectives; a single reward function is not adequate to represent such preferences. This paper proposes a method for inferring multi-objective reward-based representations of an agent's observed preferences. We model the agent's priorities over different objectives as entering \textit{lexicographically}, so that objectives with lower priorities matter only when the agent is indifferent with respect to objectives with higher priorities. We offer two example applications in healthcare---one inspired by cancer treatment, the other inspired by organ transplantation---to illustrate how the lexicographically-ordered rewards we learn can provide a better understanding of a decision-maker's preferences and help improve policies when used in reinforcement learning.}
\end{abstract}

\section{Introduction}

Modeling the preferences of agents over a set of alternatives plays an important role in many areas, including economics, marketing \citep{singh2005modeling}, politics \citep{brauninger2016modeling}, and healthcare \citep{muhlbacher2016choice}. One common approach to modeling preferences is to find a {\em utility function} over alternatives with the property that alternatives with higher utility are preferred over the ones with lower utility. This approach has been studied extensively in the machine learning (ML) literature as well---although the ML literature uses the term {\em reward function} rather than utility function.  In reinforcement learning particularly, inferring reward function from the observed behavior of an agent (viz.\ inverse reinforcement learning) has proved an effective method of replicating their policy \citep{ng2000algorithms,abbeel2004apprenticeship}. Moreover, as \citet{brown2019extrapolating,brown2019better} have recently shown, if reward functions are inferred from the preferences of an expert instead, then policies can even be improved.

In many circumstances, agent behavior is based on multiple---often competing---objectives. Healthcare in particular is replete with such circumstances. In treating cancer (and many other diseases), clinicians aim for treatment that is the most effective but also has the fewest harmful side-effects. This is especially true in radiation therapy \citep{wilkens2007imrt,jee2007lexicographic}, where  high doses are needed to be effective against tumors but also cause damage to surrounding tissue. In organ transplantation, clinicians aim to make the best match between the organ and the patient but also to give priority to patients who have been waiting the longest and/or have the most urgent need \citep{coombes2005development,schaubel2009survival}. In the allocation of resources in a pandemic, clinicians hope to minimize the spread of infection but also to safeguard the most vulnerable populations \citep{koyuncu2010optimal,gutjahr2016multicriteria}. In these situations, and many others, the preferences of decision-makers reflect the priorities they assign to various criteria.

\begin{figure}
    \includegraphics[width=\linewidth]{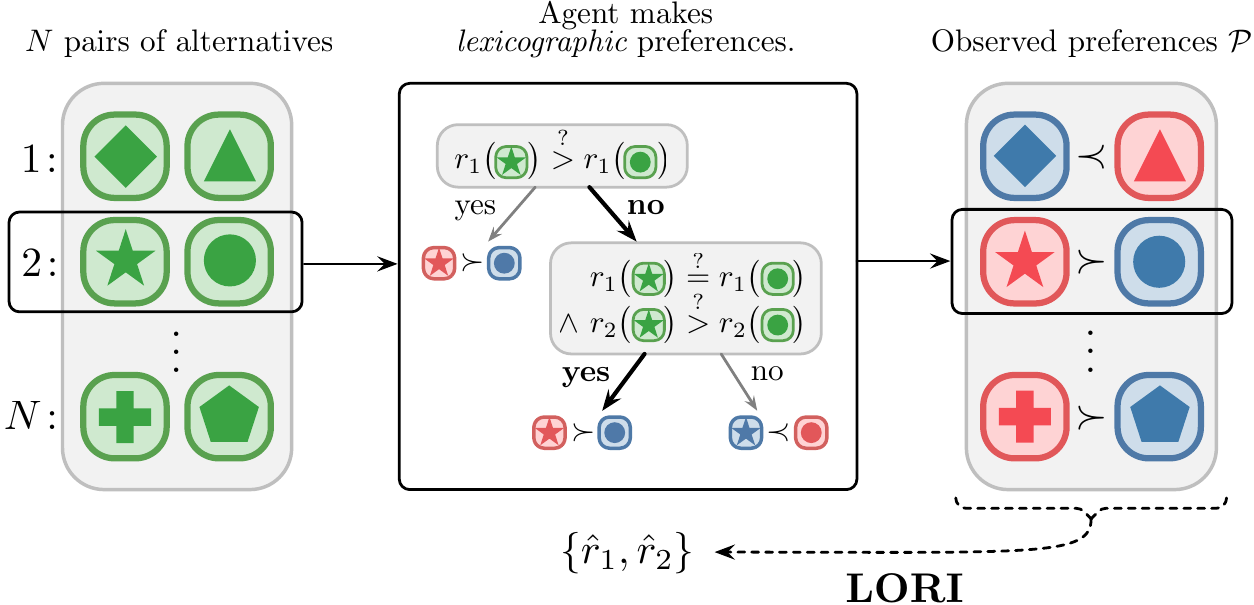}
    \caption{\textit{Setting of LORI.} We model the preferences of agents lexicographically through reward functions $r_1,r_2,...,r_k$ while allowing for errors and indifference to small differences. LORI aims to infer these reward functions from a dataset of observed preferences over pairs of alternatives.}
\end{figure}

\begin{table*}
    \centering
	\small
    \begin{tabular}{@{}*4{l}@{}}
        \toprule
        \bf Method & \bf Prototype & \bf Given & \bf Inferred \\
        \midrule
		Ordinal PL & \citet{boutilier2004cpnets} & $\mathcal{P}$ & $\succ$ \\
		Ordinal lexicographic PL & \citet{yaman2008democratic} & $\mathcal{P}$, \textit{Unordered} $\{\phi_i\in\mathbb{R}^{X}\}$ & \textit{Lex.-ordered} $\{\phi_{i'}\in\mathbb{R}^X\}$ \\
		Cardinal single-dimensional PL & \citet{chu2005preference} & $\mathcal{P}$ & $r\in\mathbb{R}^X$ \\
		\midrule
		Conventional IRL & \citet{ziebart2008maximum} & $\mathcal{D}$ & $r\in\mathbb{R}^{S\times A}$ \\
		Preference-based IRL & \citet{brown2019extrapolating} & $\mathcal{P}$ & $r\in\mathbb{R}^{S\times A}$ \\
	 	IRL w/ specifications & \citet{vazquez2018learning} & $\mathcal{D}$ & $r\in\{0,1\}^{H}$ \\
		IRL w/ constraints & \citet{scobee2020maximum} & $\mathcal{D}$, $r_2\in\mathbb{R}^{S\times A}$ & $r_1\in\{0,1\}^{S\times A}$ \\
		\midrule
        \bf LORI & \bf Ours & $\mathcal{P}$ & \textit{Lex.-ordered} $\{r_i\in\mathbb{R}^X\}$ \\
        \bottomrule
    \end{tabular}
    \caption{\rev{\textit{Comparison with related work.}} LORI is the only method that can infer lexicographically-ordered (lex.-ordered) reward functions solely from observed preferences. Preferences, demonstrations, features/attributes, and reward/utility functions are denoted by $\mathcal{P}$, $\mathcal{D}$, $\{\phi_i\}$ and $r$; the space of alternatives, states, actions, and state-action histories are denoted by $X$, $S$, $A$, and $H$.}
    \label{tbl:related}
\end{table*}

This paper provides a method for inferring multi-objective reward-based representations of a decision-maker's observed preferences. Such representations provide a better understanding of the decision-maker's preferences and promote reinforcement learning. We model priorities over different objectives as entering {\em lexicographically}, so that objectives that have lower priority matter only when the decision-maker is indifferent with respect to objective that have higher priority. \rev{While lexicographic ordering is certainly not the only way an agent might prioritize different objectives, it is a prevalent one: there is plenty of evidence showing that humans use lexicographic reasoning when making decisions \citep{kohli2007representation,slovic1975choice,tversky1988contingent,colman1999singleton,drolet2004cognitive,yee2007greedoid}.} In modeling priorities lexicographically, we take into account that the decision-maker may be indifferent to ``small'' differences. We allow for the possibility that the decision-maker may be an expert consultant, a treating clinician, the patient, or a population of experts, clinicians or patients. As we shall see, these considerations shape our model.

\paragraph{Contributions}
We introduce a new stochastic preference model based on multiple lexicographically-ordered reward functions. We formulate Lexicographically-Ordered Reward Inference (LORI) as the problem of identifying such models from observed preferences of an agent and provide a Bayesian algorithm to do so. We offer two examples---one inspired by cancer treatment, the other inspired by organ transplantation---to illustrate how the lexicographically-ordered reward functions  we learn can be used to interpret and understand the preferences of a decision-maker, and demonstrate how inferring lexicographically-ordered reward functions from preferences of an expert can help improve policies when used in reinforcement learning.

\section{Related Work}

As a method for learning reward-based representations of lexicographic preferences, LORI is related to \textit{preference learning} (PL), and as a tool for discovering lexicographically-ordered reward functions for reinforcement learning purposes, it is related to \textit{inverse reinforcement learning} (IRL).

\paragraph{Preference Learning}
Preference learning is the problem of finding a model that best explains a given set of observed preferences over a set of alternatives. It can be tackled either with an \textit{ordinal} approach, where a binary preference relation between alternatives is learned directly \citep[e.g.][]{boutilier2004cpnets}, or with a \textit{cardinal/numerical} approach, where such relations are induced through reward functions \citep[e.g.][]{chu2005preference,brochu2008active,bonilla2010gaussian,salvatore2013robust}.

Lexicographic preferences have been primarily considered from an ordinal perspective. \rev{\citet{schmitt2006complexity,dombi2007learning,yaman2008democratic,kohli2019randomized}} assume that each alternative has an unordered set of attributes (i.e. features in ML literature) and preferences are made by comparing those attributes lexicographically. They aim to learn in what order the attributes are compared. \rev{\citet{kohli2007representation,jedidi2008inferring}} consider variations of this approach including binary-satisficing lexicographic preferences, which allows indifference when comparing attributes and relaxes the assumption that attributes are compared one at a time. \rev{\citet{booth2010learning,brauning2012learning,liu2015learning,fernandes2016learning,brauning2017lexicographic,fargier2018learning}} generalize this framework and learn lexicographic preference trees instead, where the priority order of attributes is not assumed to be static but allowed to change dynamically depending on the outcome of pairwise comparisons in higher-priority attributes. All of these methods are purely ordinal rather than cardinal.

\rev{We take a different approach to modeling lexicographic preferences. In our model, each alternative has an associated multi-dimensional reward and a preference relation over alternatives is induced by the standard lexicographic preference relation between their associated rewards. The goal of LORI is to infer the reward functions that determine the reward vector of each alternative. Remember that existing methods for inferring lexicographic relations assume the set of relevant attributes to be specified beforehand (except for their priority order, which need to be inferrred). Inferring the latent reward functions in our model can be conceptualized as learning those attributes from scratch.}

\paragraph{Inverse Reinforcement Learning}
Given the demonstrated behavior of an agent, IRL aims to find a reward function that makes the demonstrated behavior appear optimal \citep{abbeel2004apprenticeship,ramachandran2007bayesian,ziebart2008maximum,boularias2011relative,levine2011nonlinear,finn2016guided}. When the demonstrated behavior is in fact optimal, the learned reward function can guide  (forward) reinforcement learning to reproduce optimal policies. However, agents do not always behave optimally according to the judgement of others---or even according to their own judgement.  In that case, conventional IRL (followed by reinforcement learning) can, at best, lead to policies that mimic the demonstrated suboptimal behavior.

Preference-based IRL is an alternative approach to conventional IRL, where a reward function is inferred from the preferences of an expert over various demonstrations instead \citep{sugiyama2012preference,wirth2016model,christiano2017deep,ibarz2018reward}. Recently, \citet{brown2019extrapolating} showed that this alternative preference-based approach can identify the intended reward function of the expert and lead to optimal policies even when the demonstrations provided are suboptimal. \citet{brown2019better} showed that these expert preferences can be generated synthetically in scenarios where it is possible to interact with the environment.

Both conventional and preference-based IRL methods focus almost exclusively on inferring a single reward function to represent preferences.  However, as we have discussed in the introduction, many important tasks are not readily evaluated in terms of a single reward function. Task representations that go beyond single reward functions has been considered in IRL. Most notably, \citet{vazquez2018learning} propose non-Markovian and Boolean specifications to describe more complex tasks, and \citet{chou2018learning,scobee2020maximum} infer the constraints that a task might have when a secondary goal is also provided. (We view tasks with constraints as the special case of lexicographically-prioritized objectives in which the reward function describing the secondary goal is only maximized when all constraints are satisfied.) To the best of our knowledge, LORI is the first reward inference method that can learn general lexicographically-ordered reward functions solely from observed preferences (as can be seen in Table~\ref{tbl:related}).

\section{Problem Formulation}

We assume that we are given observations about the preferences of a decision-maker or a set of decision-makers in the form of a pair $\mathcal{P}=(X,n)$, where $X$ is a set of alternatives and $n:X\times X\to\mathbb{Z}_+$ is a function. We interpret $n(x^{\star},x^{\circ})$ as the {\em number of times} that alternative~$x^{\star}$ was observed to be preferred to alternative~$x^{\circ}$. Note that $n(x^{\star},x^{\circ})$ may be zero because $x^{\star}$ was {\em never} observed to be preferred to $x^{\circ}$. We allow for the possibility that both $n(x^{\star},x^{\circ})>0$ and $n(x^{\circ},x^{\star})>0$, either because we are observing the preferences of a single decision-maker who is not completely consistent or because we are observing the preferences of a {\em population} of decision-makers who do not entirely agree. Write $A=\{(x^{\star},x^{\circ})\in X\times X: n(x^{\star},n^{\circ})>0\}$; this is the set of pairs $(x^{\star},x^{\circ})$ for which $x^{\star}$ is observed to be preferred to $x^{\circ}$ at least once. If $(x^{\star},x^{\circ})\in A$, we often write $x^{\star}\succ x^{\circ}$. (Note that we {\em do not} assume that $\succ$ is a preference relation in the sense used in economics; in particular, we {\em do not} assume that the preference relation is asymmetric or transitive.)

We seek to explain these observations in terms of an ordered set of $k$ reward functions~$\{r_i\in\mathbb{R}^X\}_{i=1}^k$ (numbered so that $r_1$ is prioritized over $r_2$, $r_2$ is prioritized over $r_3$, and so on), in the sense that $x^{\star}\succ x^{\circ}$ tends to be observed if the reward vector~$\bm{r}(x^{\star})=\{r_1(x^{\star}),\ldots,r_k(x^{\star})\}$ lexicographically dominates the reward vector~$\bm{r}(x^{\circ})$, in which case we write $\bm{r}(x^{\star})>_{\mathrm{lex}}\bm{r}(x^{\circ})$. \rev{Formally, $\bm{r}(x^{\star})>_{\mathrm{lex}}\bm{r}(x^{\circ})$ if and only if there exists $i\in\{1,\ldots,k\}$ such that $r_i(x^{\star})>r_i(x^{\circ})$ and $r_j(x^{\star})=r_j(x^{\circ})$ for all $j<i$.} Because we allow for the possibility that $x^{\star}\succ x^{\circ}$ and $x^{\circ}\succ x^{\star}$, we incorporate randomness; we also allow for indifference in the presence of small differences. Our objective can be summarized as:

\paragraph{Objective}
Infer reward functions~$\{r_i\}_{i=1}^k$ from the observed preferences~$\mathcal{P}$ of a decision-maker.
It should be emphasized that LORI \textit{does not} assume that there are reward functions and a lexicographic ordering that represent the given preferences \textit{exactly}; it simply attempts to find reward functions and a lexicographic ordering that represents the given preferences \textit{as accurately as possible}.

\begin{figure*}
    \centering
    \begin{subfigure}{.33\linewidth}
        \centering
        \includegraphics[width=.9\linewidth]{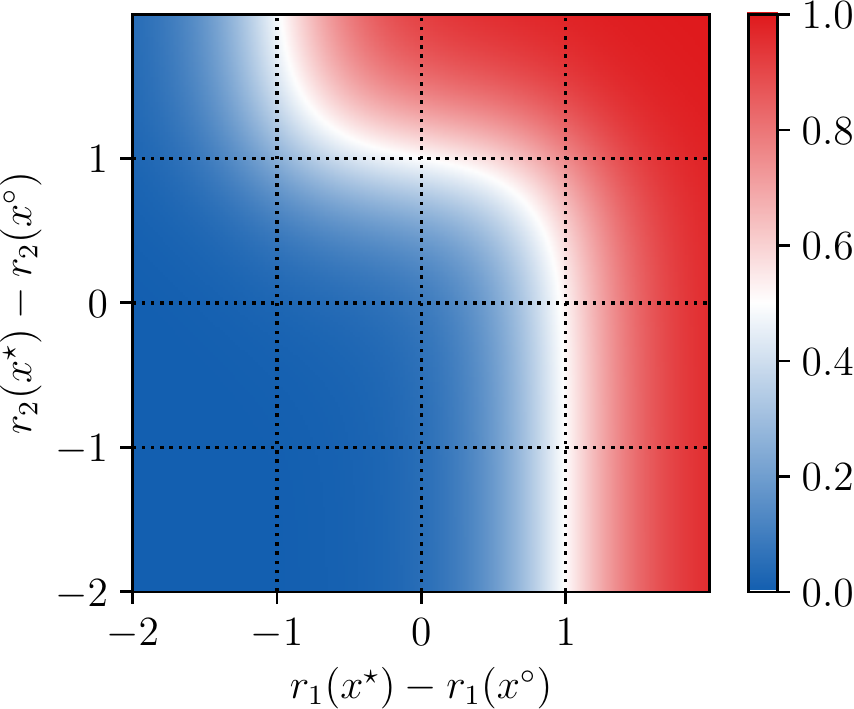}
        \caption{\textbf{Baseline preferences}\\($\alpha_1=\alpha_2=3$, $\varepsilon_1=\varepsilon_2=1$)}
    \end{subfigure}%
    \begin{subfigure}{.33\linewidth}
        \centering
        \includegraphics[width=.9\linewidth]{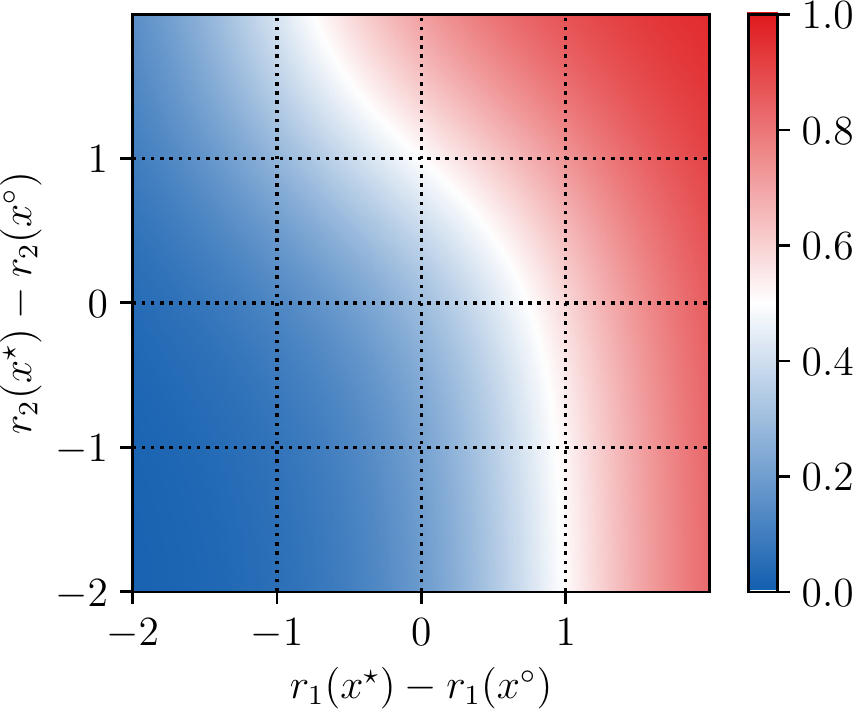}
        \caption{\textbf{Less consistent preferences}\\($\alpha_1=\alpha_2=1.5$, $\varepsilon_1=\varepsilon_2=1$)}
    \end{subfigure}%
    \begin{subfigure}{.33\linewidth}
        \centering
        \includegraphics[width=.9\linewidth]{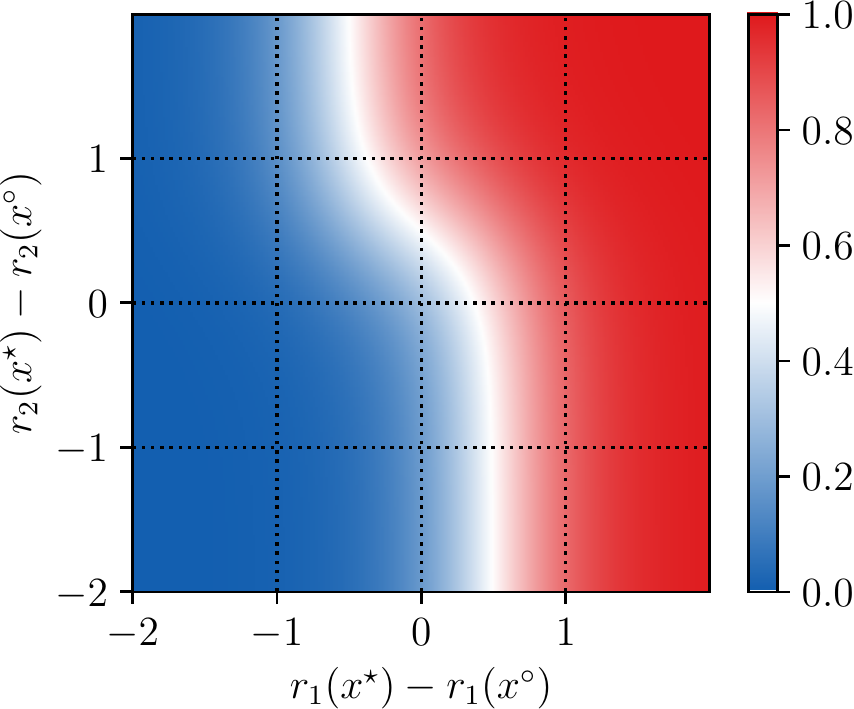}
        \caption{\textbf{Less tolerant preferences}\\($\alpha_1=\alpha_2=3$, $\varepsilon_1=\varepsilon_2=0.5$)}
    \end{subfigure}
    \caption{\textit{Probability distribution $\Pr(x^{\star}\succ x^{\circ}|r_1,r_2)$ over preferences induced by reward functions~$r_1$, $r_2$.} Smaller $\alpha_i$'s lead to preferences that are less consistent while smaller $\varepsilon_i$'s lead to ones less tolerant of differences.} \label{fig:model}
    \vspace{-\baselineskip}
\end{figure*}

\section{Lexicographically-ordered Reward Inference}

A substantial literature views $x^{\star}\succ x^{\circ}$ as a {\em random event} and models the probability of this event in terms of a noisy comparison between the rewards of $x^{\star}$ and $x^{\circ}$. This idea is the foundation of the {\em logistic choice model} originated by \citet{mcfadden1973conditional}; recent ML work includes \citet{brown2019extrapolating,brown2019better}. Formally, given a reward function $r\in\mathbb{R}^X$ and a parameter $\alpha\geq0$, this literature defines
\begin{equation}
	\begin{aligned}
		\Pr(x^{\star}\succ x^{\circ}|r) &= \frac{e^{\alpha r(x^{\star})}}{e^{\alpha r(x^{\star})}+e^{\alpha r(x^{\circ})}} \\ &= \frac{1}{1+e^{-\alpha(r(x^{\star})-r(x^{\circ}))}} ~. \label{eqn:model-single}
	\end{aligned}
\end{equation}
The parameter $\alpha$ represents the extent to which preference is {\em random}. (Randomness of the preference may reflect inconsistency on the part of the decision-maker, or reflect the aggregate preferences of some population of decision-makers---experts or clinicians or patients.) Note that if $\alpha=0$ then preference is uniformly random; at the opposite extreme, as $\alpha\to\infty$ then preference becomes perfectly deterministic.

Given a reward function $r$, parameter $\alpha$, observed preferences $\mathcal{P} = (X,n)$ and alternatives
$x^\star,x^\circ \in X$, we can ask how likely it is that $r, \alpha$ would generate the same relationship between $x^\star,x^\circ$ that we observe in $\mathcal{P}$. By definition, $n(x^\star,x^\circ)$ is the number of times that $x^\star$ is observed to be preferred to $x^\circ$ and $n(x^\circ,x^\star)$ is the number of times that $x^\circ$ is observed to be preferred to $x^\star$, so  $N(x^\star,x^\circ)= n(x^\star,x^\circ) + n(x^\circ,x^\star)$ is the total number of times that $x^\star, x^\circ$ are observed to be compared. The probability that the preference generated by $r$ agrees with the observations in $\mathcal{P}$ with respect to $x^\star,x^\circ$ is
\begin{equation}
	\begin{aligned}
		\Pr(\mathcal{P}_{x^{\star},x^{\circ}}|r) &= {\textstyle{N(x^\star,x^\circ)\choose n(x^\star,x^\circ)}} \\
    	&\hspace{12pt}\times \Pr(x^{\star}\succ x^{\circ}|r)^{n(x^\star,x^\circ)} \\
		&\hspace{12pt}\times \Pr(x^{\circ}\succ x^{\star}|r)^{n(x^\circ,x^\star)} \makebox[0pt][l]{~.}
	\end{aligned}
\end{equation}
Hence, the probability that the preference generated by $r$ agrees with $\mathcal{P}$ (i.e.\ the likelihood of $r$ with respect to $\mathcal{P}$) is
\begin{equation}
	\begin{aligned}
    	\mathcal{L}(r;\mathcal{P}) &= \Pr(\mathcal{P}|r) \\
		&= \big[{\textstyle\prod_{(x^{\star},x^{\circ})\in X\times X}}\Pr(\mathcal{P}_{x^\star,x^\circ}|r)\smash{\big]^{1/2}} ~. \label{eqn:likelihood-single}
	\end{aligned}
\end{equation}
(The exponent $1/2$ is needed because each of the terms $\Pr(\mathcal{P}_{x^{\star},x^{\circ}}|r)$ appears twice: once indexed by $(x^{\star},x^{\circ})$ and again indexed by $(x^{\circ},x^{\star})$.) If the ``true'' reward function is known/assumed to belong to a given family~$\mathcal{R}$, it can be inferred by finding the maximum likelihood estimate (MLE) $\hat{r}=\argmax_{r\in \mathcal{R}}\mathcal{L}(r;\mathcal{P})$.

\paragraph{Lexicographic Setting}
Adapting this probabilistic viewpoint to our context requires two changes. The first is that we make probabilistic comparisons for {\em multiple} reward functions. The second is that we allow for the possibility that the rewards $r(x^{\star}), r(x^{\circ})$ may not be {\em exactly} equal but the difference between them may be regarded as negligible. We therefore begin by defining:
\begin{align}
    \Pr(x^{\star}\succ_i x^{\circ} | r_i) &= \frac{1}{1+e^{-\alpha_i(r_i(x^{\star})-r_i(x^{\circ})-\varepsilon_i)}} ~, \nonumber \\
    \Pr(x^{\star}\prec_i x^{\circ} | r_i) &= \frac{1}{1+e^{-\alpha_i(r_i(x^{\circ})-r_i(x^{\star})-\varepsilon_i)}} ~, \\[6pt]
    \Pr(x^{\star}\equiv_i x^{\circ} | r_i) &= 1\!-\!\Pr(x^{\star}\!\succ_i\! x^{\circ}|r_i)\!-\!\Pr(x^{\star}\!\prec_i\! x^{\circ}|r_i) ~. \nonumber
\end{align}
Respectively, these are the probabilities that $x^{\star}$ is regarded as significantly better than, significantly worse than or not significantly different from $x^{\circ}$ when measured in terms of the reward function~$r_i$. As before, the parameter~$\alpha_i\geq 0$ represents the extent to which the reward is random. The parameter~$\varepsilon_i\geq 0$ measures the extent to which the rewards of $x^{\star}$ and $x^{\circ}$ must differ for the difference to be regarded as ``significant.''  Our model is then given by
\begin{equation}
	\begin{aligned}
    &\Pr(x^{\star}\succ x^{\circ}|\{r_i\}_{i=1}^k) \\
	&\hspace{12pt}= \sum_{i=1}^k\bigg[ \Pr(x^{\star}\succ_i x^{\circ}|r_i)\cdot\prod_{j=1}^{i-1}\Pr(x^{\star}\equiv_j x^{\circ}|r_j) \bigg] ~. \label{eqn:model}
	\end{aligned}
\end{equation}
Figure~\ref{fig:model} offers a visual description of how the parameters $\alpha_i$ and $\varepsilon_i$ control the properties of this preference distribution.

As before, the probability that the preference generated by $\{r_i\}$ agrees with $\mathcal{P}$ with respect to $x^{\star},x^{\circ}$ is
\begin{equation}
	\begin{aligned}
		\Pr(\mathcal{P}_{x^{\star},x^{\circ}}|\{r_i\}) &= {\textstyle{N(x^\star,x^\circ)\choose n(x^\star,x^\circ)}} \\
		&\hspace{12pt}\times \Pr(x^{\star}\succ x^{\circ}|\{r_i\})^{n(x^\star,x^\circ)} \\
		&\hspace{12pt}\times \Pr(x^{\circ}\succ x^{\star}|\{r_i\})^{n(x^\circ,x^\star)} \makebox[0pt][l]{~.}
	\end{aligned}
\end{equation}
Hence, the probability that the preference generated by $\{r_i\}$ agrees with $\mathcal{P}$ (i.e.\ the likelihood of $\{r_i\}$ w.r.t.\ $\mathcal{P}$) is
\begin{equation}
	\begin{aligned}
		\mathcal{L}(\{r_i\};\mathcal{P}) &= \Pr(\mathcal{P}|\{r_i\}) \\
		&= \big[{\textstyle\prod_{(x^{\star},x^{\circ})\in X\times X}}\Pr(\mathcal{P}_{x^\star,x^\circ}|\{r_i\})\smash{\big]^{1/2}} ~. \label{eqn:likelihood}
	\end{aligned}
\end{equation}
And, as before, if the ``true'' reward functions are known/assumed to belong to a given family~$\mathcal{R}$, they can be inferred by finding the maximum likelihood estimate $\{\hat{r}_i\}=\argmax_{\{r_i\in \mathcal{R}\}}\mathcal{L}(\{r_i\};\mathcal{P})$.

\rev{Now, suppose we consider a parameterized family of reward functions such that $r_i=r_{\theta_i}$ for $\theta_i\in\Theta$, where $\Theta$ is the space of parameters. Then, the (approximate) MLE of the parameters~$\{\theta_i\}_{i=1}^k$ can simply be found via gradient descent using the negative log-likelihood~$\lambda=-\log\mathcal{L}(\{r_{\theta_i}\};\mathcal{P})$ as the loss function. This is because $\lambda$ is differentiable with respect to $\{\theta_i\}$ as long as $r_{\theta}$ is a differentiable parameterization with respect to $\theta$ (which we show in \com{Appendix~A}). In our exposition so far, we have chosen to treat $\alpha_i$'s and $\varepsilon_i$'s as hyperparameters for simplicity. However, in practice, they can easily be treated as free variables and learned alongside with parameters~$\{\theta_i\}$, which we will be doing in all of our experiments. (We show that $\lambda$ is differentiable with respect to $\{\alpha_i\}$ and $\{\varepsilon_i\}$ in \com{Appendix~A} as well.)}

\rev{Finally, it needs to be emphasized that LORI is inherently capable of identifying the priority order of reward functions as well as which reward functions are relevant to modeling preferences. This is because different permutations of a given set of parameters~$\{\theta_i\}_{i=1}^k$ are all present in our search space~$\Theta^k$. In contrast, ordinal models of lexicographic preferences assume relevant attributes to be specified beforehand and only learn the priority order of those attributes.}

\subsection{Analysis of LORI}

Our model satisfies the following desirable properties:
\begin{enumerate}[label=(\roman*)]
    \item Setting $k=1$ and $\varepsilon_1=0$ reproduces the logistic model given in \eqref{eqn:model-single},
    \item Taking $\alpha_i\to\infty$, $\varepsilon_i\to0$, and $\alpha_i\varepsilon_i\to\infty$ yields the no-errors, deterministic case,
    \item The parameters $\varepsilon_i$ have natural interpretations. They are the thresholds above which a reward difference is considered significant: $\Pr(x^{\star}\succ_i x^{\circ}|r_i)>1/2$ if and only if $r_i(x^{\star})-r_i(x^{\circ})>\varepsilon_i$.
\end{enumerate}

It is worth noting that lexicographic representations using {\em multiple} reward functions are not only convenient, but (often) {\em necessary}.  For the simplest example, consider the ordinary lexicographic preference relation on $\mathbb{R}^2$: $(x_1,x_2)\succ(x'_1,x'_2)$ if $x_1>x'_1$ or $x_1=x'_1$ and $x_2>x'_2$. It is well-known that there does not exist any reward function $r: \mathbb{R}^2\to\mathbb{R}$ that represents $\succ$. That is, there is no reward function $r$ with the property that $(x_1,x_2)\succ(x'_1,x'_2)$ if and only if $r(x_1,x_2)>r(x'_1,x'_2)$. Similarly, the ordinary lexicographic ordering $\succ$ on $\mathbb{R}^{k+1}$ cannot be represented by a lexicographic ordering involving only $k$ reward functions.

When we consider probabilistic orderings and allow for errors, we can again find simple situations in which preferences that employ $k+1$ reward functions cannot be represented in terms of $k$ reward functions.  For example, consider $X=\mathbb{R}^2$; let the reward functions $r_1,r_2:\mathbb{R}^2\to\mathbb{R}$ be the coordinate projections and take $\alpha_1=\alpha_2=1$ and $\varepsilon_1=\varepsilon_2=1$. The probabilistic preference relation $\succ$ defined by these reward functions and parameters is {\em intransitive}. For example, consider the points $x,x',x''\in\mathbb{R}^2$ defined by: $x=(-0.6, 2)$, $x'=(0,0)$, $x''=(0.6,-2)$. Direct calculation shows that $\Pr(x\!\succ\! x')>0.5$, $\Pr(x' \!\succ\! x'')>0.5$, and $\Pr(x''\!\succ\! x)>0.5$. On the other hand, suppose we are given a reward function $r:\mathbb{R}^2\to\mathbb{R}$, and parameters $\alpha\geq 0$ and $\varepsilon\geq 0$. If we define $\succ'$ to be the probabilistic preference relation defined by $r,\alpha,\varepsilon$, then $\succ'$ is necessarily {\em transitive}. To see this, note in order that if $\Pr(x^{\star}\succ'x^{\circ})>0.5$, we must have $\alpha(r(x^\star)-r(x^\circ)-\varepsilon)>0$; in particular, we must have $r(x^{\star})-r(x^\circ)>\varepsilon$. Hence, if $\Pr(x\succ'x')>0.5$ and $\Pr(x'\succ'x'')>0.5$ then we must have $r(x)-r(x')>\varepsilon$ and $r(x')-r(x'')> \varepsilon$, whence $r(x)-r(x'')>2\varepsilon>\varepsilon$ and $\Pr(x\succ'x'')>0.5$.

\rev{Finally, a reasonable question to ask is how to determine the number of reward functions~$k$ when using LORI. For lexicographic models, even when the number of potential criteria considered by the agent is large, the number of criteria that actually enter into preference is likely to remain small. Remember that, in a lexicographic model, the $i$-th most important criterion will only be relevant for a particular decision if the $i-1$ more important criteria are all deemed equivalent. For most decision, this would be unlikely for even moderately large $i$. This means using a lexicographic model that employs only a few criteria (i.e.\ a model with small $k$) would be enough in most cases; increasing $k$ any further would have little to no benefit in terms of accuracy. We demonstrate this empirically in \com{Appendix~B}.}

\section{Illustrative Examples in Healthcare}

Inferring lexicographically-ordered reward functions from preferences can be used either to improve decision-making behavior or to understand it. Here, we give examples of each in healthcare. However, it is worth noting that applications of LORI are not limited to the healthcare setting; it can be applied in any decision-making environment where multiple objectives are at play.

\subsection{Improving Behavior: Cancer Treatment}

Consider the problem of treatment planning for cancer patients. For a given patient, write $a_t\in A=\{0,1\}$ for the binary application of a treatment such as chemotherapy, $z_t\in Z=\mathbb{R}$ for tumor volume (as a measure of the treatment's efficacy), and  $w_t\in W=\mathbb{R}$ for the white blood cell (WBC) count (as a measure of the treatment's toxicity) at time $t\in\{1,2,\ldots\}$. In our experiments, we will simulate the tumor volume~$z_t$ and the WBC count~$w_t$ of a patient given a treatment plan~$a_t$ according to the pharmacodynamic models of \citet{iliadis2000optimizing}:
\begin{equation}
    \begin{aligned}
        z_{t+1} &= z_t + 0.003 z_t\log(1000/z_t) - 0.15 z_ta_t + \nu_t \\
        w_{t+1} &= w_t + 1.2 - 0.15 w_t - 0.4 w_ta_t + \eta_t ~,
    \end{aligned}
\end{equation}
where $\nu_t\sim\mathcal{N}(0,0.5^2)$ and $\eta_t\sim\mathcal{N}(0,0.5^2)$ are noise variables, $z_1\sim\mathcal{N}(30,5^2)$, and $w_1=8$. Notice that both the tumor volume and the WBC count decrease when the treatment is applied and increase otherwise. We  assume that clinicians aim to minimize the tumor volume while keeping the average WBC count above a threshold of five throughout the treatment. We define the set of alternatives to be all possible treatment trajectories: $X=\bigcup_{\tau=1}^{\infty}(A\times Z\times W)^{\tau}$.  Then, the treatment objective can be represented in terms of lexicographic reward functions:
\begin{equation}
	\begin{aligned}
		r_1(\bm{a}_{1:\tau},\bm{z}_{1:\tau},\bm{w}_{1:\tau}) &= \min\big\{5,{\textstyle\frac{1}{\tau}}{\textstyle\sum_{t=1}^{\tau}}w_t\big\} ~, \\
		r_2(\bm{a}_{1:\tau},\bm{z}_{1:\tau},\bm{w}_{1:\tau})&= -{\textstyle\frac{1}{\tau}}{\textstyle\sum_{t=1}^{\tau}}z_t ~.
	\end{aligned} \label{eqn:cancer-rewards}
\end{equation}

A  {\em policy} $\pi\in\Delta(A)^{Z\times W}$ is a function from the features of a patient at time~$t$ to a distribution over actions such that $a_t\sim\pi(\cdot|z_t,w_t)$. By definition, the optimal policy is given by $\pi^*=\argmax_{\mathrm{lex},\pi}\Ex[\bm{r}(\bm{a}_{1:\tau},\bm{z}_{1:\tau},\bm{w}_{1:\tau})]$. We {\em do not} assume that clinicians follow the optimal policy, but rather follow some policy that approximates the optimal policy: $\pi_b(a|\bm{z}_{1:t},\bm{w}_{1:t})=(1-\epsilon)\pi^*(a|\bm{z}_{1:t},\bm{w}_{1:t})+\epsilon/|A|$ for some $\epsilon>0$, which we will call the behavior policy. This means that the policy actually followed by clinicians is improvable.

Now, suppose we want to improve the policy followed by clinicians using reinforcement learning but we do not have access to their preferences explicitly in the form of reward functions~$\{r_1,r_2\}$ so we cannot compute the optimal policy~$\pi^*$ directly. Instead, we have access to some demonstrations~$\mathcal{D}\subset X$ generated by clinicians as they follow the behavior policy
$\pi_b$ and we  query an expert or panel of experts (which might consist of the clinicians themselves) over the demonstrations/alternatives in $\mathcal{D}$ to obtain a set of observed preferences~$\mathcal{P}$.  We use LORI to estimate the reward functions~$\{r_1,r_2\}$ from the observed preferences~$\mathcal{P}$.

\paragraph{Setup}
For each experiment, we take $\epsilon=0.5$ and generate $1000$ trajectories with $\tau=20$ to form the demonstration set~$\mathcal{D}$. Then, we generate preferences by sampling $1000$ trajectory pairs from $\mathcal{D}$ and evaluating according  to the ground-truth reward functions~$\{r_1,r_2\}$ and the model given in \eqref{eqn:model}, where
$\varepsilon_1=\varepsilon_2=0.1$ and $\alpha_1=\alpha_2=10 \log(9)$ (ties are broken uniformly at random). These form the set of expert preferences~$\mathcal{P}$.

We infer $k=2$ reward functions from the expert preferences~$\mathcal{P}$ using LORI; for comparison, we infer a single reward function using T-REX \citep{brown2019extrapolating}, which is the single-dimensional counterpart of LORI (with $k=1$), and another single reward function from demonstrations~$\mathcal{D}$ instead of preferences~$\mathcal{P}$ using Bayesian IRL \citep{ramachandran2007bayesian}. (Keep in mind that LORI infers lexicographic reward functions but the two alternatives infer only a single reward function.) LORI also infers parameters~$\varepsilon_1$ and $\varepsilon_2$ together with $r_1,r_2$. (We set $\alpha_1=\alpha_2=1$; there is no loss of generality because the other variables simply scale.)

Then, using the estimated reward functions, we train optimal policies using reinforcement learning. In the case of LORI, we use the algorithm proposed in \citet{gabor1998multi}, which is capable of optimizing thresholded lexicographic objectives. We also learn a policy directly from the demonstration set~$\mathcal{D}$ by performing behavioral cloning (BC), where actions are simply regressed on patient features. The resulting policy aims to mimic the behavior policy~$\pi_b$ as closely as possible. Details regarding the implementation of algorithms can be found in \com{Appendix C}. We repeat each experiment five times to obtain error bars.

\begin{table}
    \centering
    \small
    \begin{tabular}{@{}l*2{c}@{}}
        \toprule
        \bf Alg. & \bf \makecell{RMSE} & \bf \makecell{Accuracy} \\
        \midrule
        BIRL & $0.323\!\pm\! 0.005$ & $88.1\%\!\pm\! 0.95\%$ \\
        T-REX & $0.214\!\pm\! 0.007$ & $89.1\%\!\pm\! 0.61\%$ \\
        LORI & $0.103\!\pm\! 0.089$ & $92.4\%\!\pm\! 2.71\%$ \\
        \bottomrule
    \end{tabular}
    \caption{\textit{Comparison of reward functions based on preference prediction performance.} LORI performs the best.}
    \label{tbl:reward}
\end{table}

\begin{table}
    \centering
    \small
    \begin{tabular}{@{}l@{\hspace{4.5pt}}|*5{@{\hspace{4.5pt}}c}@{}}
        \toprule
        & \bf Behavior & \bf BC & \bf BIRL & \bf T-REX & \bf LORI \\
        \midrule
        BC & $55.7\!\pm\! 1.3$ & -- \\
        BIRL & $65.6\!\pm\! 1.1$ & $57.3\!\pm\! 2.1$ & -- \\
        T-REX & $65.7\!\pm\! 1.2$ & $57.5\!\pm\! 2.3$ & $49.8\!\pm\! 0.9$ & -- \\
        LORI & $71.5\!\pm\! 3.8$ & $65.6\!\pm\! 5.0$ & $69.4\!\pm\! 15$ & $68.9\!\pm\! 15$ & -- \\
        Optimal & $75.1\!\pm\! 1.3$ & $70.0\!\pm\! 1.6$ & $82.5\!\pm\! 0.2$ & $82.4\!\pm\! 0.5$ & $62.0\!\pm\! 16$ \\
        \bottomrule
    \end{tabular}
    \caption{\textit{Comparison of policies based on how often the row policies are preferred to the column policies.} Note that reported values represent percentages. LORI is the most frequently preferred to the behavior policy and its performance is comparable to that of the optimal policy.}
    \label{tbl:policy}
\end{table}


\paragraph{Results}
To evaluate the quality of reward functions learned by LORI, T-REX, and BIRL, we compare the prediction accuracy of the preference models they induce on a test set of $1000$ additional preferences; the results are shown in Table~\ref{tbl:reward}.  We see that LORI performs better than either T-REX and BIRL. This is because LORI is the only method to capture the multi-objective nature of the clinicians' goal (and the expert's preferences). BIRL performs the worst since the demonstrations in $\mathcal{D}$ are suboptimal.

Table~\ref{tbl:policy} compares the performance of various policies: the behavior policy, the policy that is learned via BC, the policies that are trained on the basis of reward functions learned by BIRL, T-REX and LORI, and the true optimal policy. Each entry in Table~\ref{tbl:policy} shows the frequency with which the row policy is preferred to the column policy. Letting $x\sim\pi$ be the distribution of trajectories~$x$ generated by following policy~$\pi$, the frequency that policy~$\pi^{\star}$ is preferred to the policy~$\pi^{\circ}$ is defined to be $\Pr(\pi^{\star}\succ\pi^{\circ})=\Ex_{x^{\star}\sim\pi^{\star},x^{\circ}\sim\pi^{\circ}}\Pr(x^{\star}\succ x^{\circ}|r_1,r_2)$. (This is  estimated by sampling $1000$ trajectories from both policies.) We use the frequency with which one policy is preferred to another as the measure for the improvement provided by the first policy over the second. (Note that because there is no single ground-truth reward function, it is not feasible to compare the \textit{values} of the two policies, which would have been the usual measure used in reinforcement learning.) As can be seen, LORI improves on other methods, improves on the behavioral policy more often than other methods, and performs comparably to the true optimal policy.

\rev{\com{Appendix~B} includes additional experiments where the ground-truth preferences are generated by a single reward function (rather than $r_1,r_2$ defined in \eqref{eqn:cancer-rewards}). Since our preference model is strictly a generalization of single-dimensional models, LORI (with $k=2$) still performs the best---but it does not improve over T-REX as much.}

\subsection{Understanding Behavior: Organ Transplantation}

Consider the organ allocation problem for transplantation. Write $P$ for the space of all patients (or patient characteristics) and $O$ for the set of organs. At a given time $t\in T$, there is a set of patients $P_t\subset P$ who are waiting for an organ transplant. When an organ~$o_t\in O$ becomes available at time~$t$, an allocation policy matches that organ with a particular patient~$p_t\in P_t$. In effect, the allocation policy expresses a preference relation: letting $X=P\times O$ be the set of alternatives, the pairing~$(p_t,o_t)\in X$ is preferred over alternative pairings~$\{(p',o_t)\}_{p'\in P_t\setminus\{p_t\}}$ that were also possible at time~$t$. From these observations, we can infer a lexicographic reward representation $\{r_i\}_{i=1}^k$ that explains the decisions made by the allocation policy.  (Notice that if we view elements of $P$ as patient characteristics, then the observed preferences need not be consistent over time.)

\paragraph{Setup}
We investigate the liver allocation preferences in the US in terms of the \textit{benefit} a transplantation could provide and the \textit{need} of patients for a transplant. These two objectives are at odds with each other since the patient in greatest need might not be the patient who would benefit the most from an available organ. As a measure of benefit, we consider the estimated additional number of days a patient would survive if they were to receive the available organ, and as a measure of need, we consider the estimated number of days they would survive without a transplant. We estimate both benefit and need via Cox models following the same methodology as TransplantBenefit, which is used in the current allocation policy of the UK \citep{neuberger2008selection}. Our analysis is based on the Organ Procurement and Transplantation Network (OPTN) data for liver transplantations as of December 4, 2020. From the OPTN data, we only consider patients who were in the waitlist during 2019 and for whom benefit and need can be estimated from the recorded features. This leaves us with $4070$ patients, $3942$ organ arrivals, and $2,\!450,\!718$ observations about the allocation policy's preferences over patient-organ pairings.

We use LORI (with $k=2$) and T-REX \citep{brown2019extrapolating}, which is the single-dimensional counterpart of LORI ($k=1$, $\varepsilon_1=0$), to infer reward functions that are linear with respect to benefit and need. (The restriction to linear reward functions means that preferences depend only on the benefit and need {\em differences} between the two pairings.) In the case of LORI, we also infer $\varepsilon_1$ and $\varepsilon_2$ to determine what amount of benefit or need is considered to be significant by the allocation policy. For T-REX, we set $\alpha=1$, and for LORI, we set $\alpha_1=\alpha_2=1$ (again without any loss of generality).


\begin{figure}
    \centering
    \includegraphics[width=.75\linewidth]{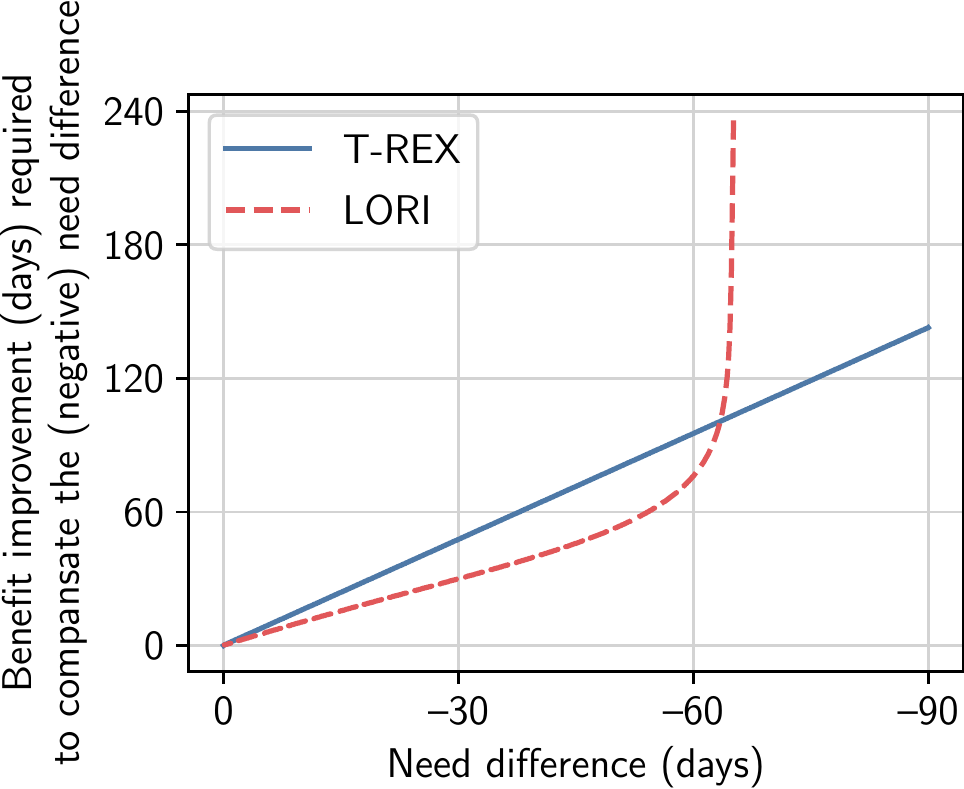}
    \caption{\textit{Trade-off between benefit and need according to T-REX and LORI.} For both T-REX and LORI, there is a trade-off of benefit against need.  For T-REX, this trade-off occurs at a constant rate.  For LORI, the trade-off occurs at an almost constant rate when the need difference is below 60 days, but above 60 days the required trade-off of benefit against need increases sharply, and above 65 days it is almost impossible for benefit to compensate for the need difference.}
    \label{fig:organ-priority}
\end{figure}

\paragraph{Results}
The single-dimensional reward function learned by T-REX is
\begin{equation}
    r = 0.0086\cdot\textit{Benefit} + 0.0137\cdot\textit{Need}
\end{equation}
while the lexicographically-ordered reward functions learned by LORI are
\begin{equation}
    \begin{aligned}
        r_1 &= 0.0001\cdot\textit{Benefit} + 0.0139\cdot\textit{Need} \\
        r_2 &= 0.0562\cdot\textit{Benefit} + 0.0002\cdot\textit{Need}    
    \end{aligned}
\end{equation}
with $\varepsilon_1=0.8944$ and $\varepsilon_2=1.8830$.

As Figure~\ref{fig:organ-priority} shows, LORI reveals that need is \textit{prioritized} over benefit.  This can be seen by the weights in the primary reward function $r_1$ and secondary reward function $r_2$, but more specifically in the finding that a need difference  greater than 65 days cannot be outweighed by {\em any} benefit difference;  {\em guaranteeing} that the patient with greater need will receive the organ.  By contrast, there is no such finding for T-REX: any need difference can be outweighed by a sufficiently large benefit difference.
This finding is  consistent with  current allocation practices in the US. When an organ becomes available for transplantation, it is offered to a patient based on their MELD score \citep{wiesner2003model}, which is strictly a measure of how sick the patient is, and so considers only the patient's need and not the benefit they will obtain. However after an offer is made, clinicians might still choose to reject the offer in order to wait for a future offer that would be more beneficial for their patient. Since offers are first made on the basis of need and then accepted/rejected on the basis of benefit, it is reasonable for need to have priority over benefit in the representation learned by LORI.

\begin{figure}[t]
    \begin{subfigure}{\linewidth}
        \centering
        \includegraphics[width=.75\linewidth]{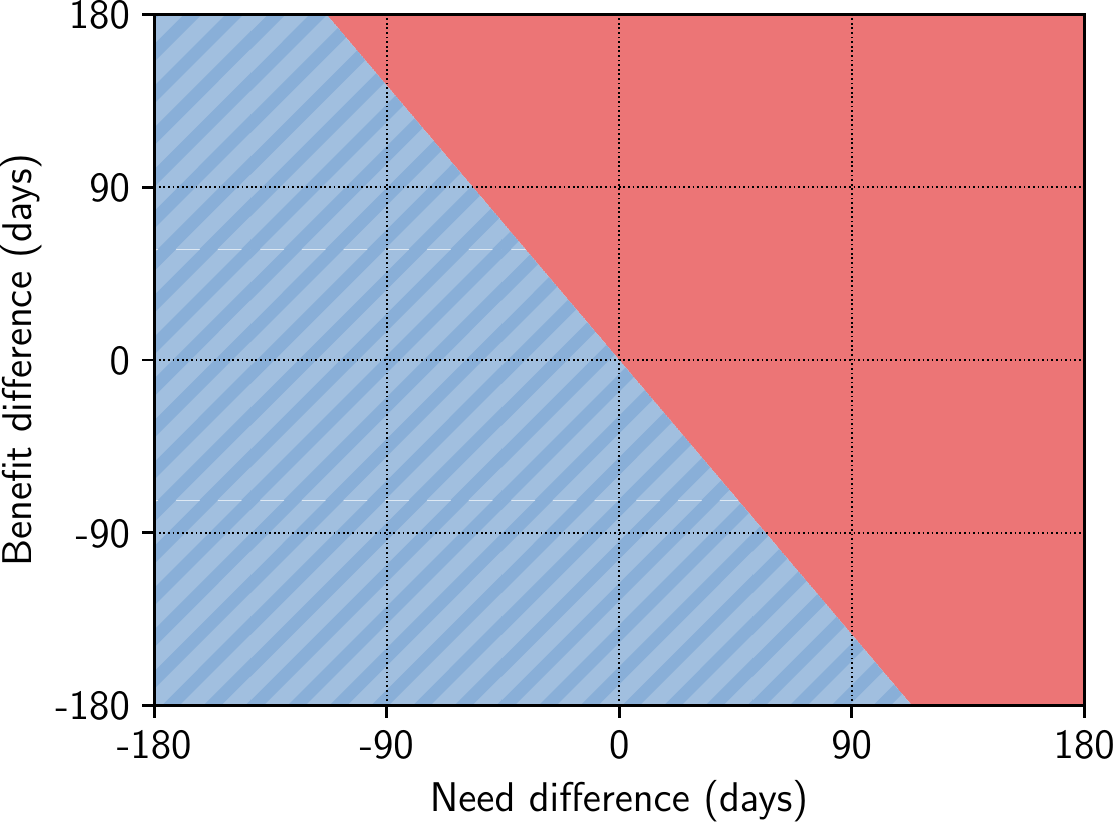}
        \caption{\textbf{Preferences learned by T-REX}}
        \vspace{1.2pt}
    \end{subfigure}
    \begin{subfigure}{\linewidth}
        \centering
        \includegraphics[width=.75\linewidth]{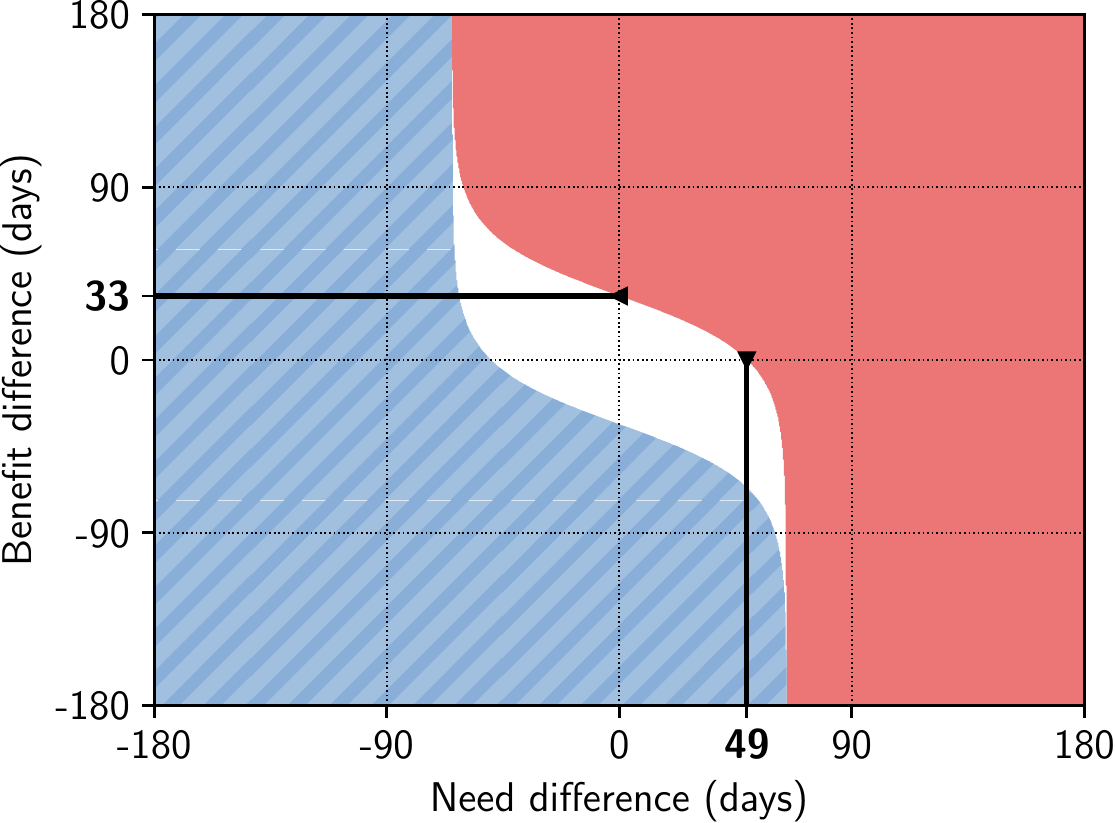}
        \caption{\textbf{Preferences learned by LORI}}
        \label{fig:organ-lori}
    \end{subfigure}
    \caption{\textit{Preferences with respect to benefit and need differences between a candidate pairing and an alternative pairing.} Red (solid) means the candidate pairing while blue (striped) means the alternative pairing is likely to be preferred with probability at least $0.5$. For LORI, white denotes a region of indifference, where the probability of preferring either pairing is less than $0.5$.}
    \label{fig:organ}
\end{figure}

Figure~\ref{fig:organ} depicts visually the preferences induced by the two alternative representations.
As pointed out above, in the preferences learned by LORI, a need difference greater than 65 days cannot be outweighed by any benefit difference, while in the preferences learned by T-REX, any need difference can be outweighed by a sufficiently large benefit difference (and the rate at which benefit trades-off for need is constant).  Moreover, LORI identifies a  region (indicated by white in Figure~\ref{fig:organ-lori}), where differences in benefit and need become insignificantly small and preferences are made mostly at random (at least with respect to benefit and need).

\section{Conclusion}

We proposed LORI, a method for inferring lexicographically-ordered reward functions from observed preferences of an agent. Through examples from healthcare, we showed that learning such reward functions can be useful in either improving or understanding behavior. While we have modeled priorities over different objectives lexicographically, which happens to be the case in many decision-making scenarios including our healthcare examples, there are numerous other ways an agent might reason about multiple objectives. Future work can focus on inferring preference representations based on alternative notions of multi-objective decision-making.

\section{Acknowledgements}

This work was supported by the US Office of Naval Research (ONR) and the National Science Foundation (NSF, grant number 1722516). Part of our experimental results are based on the liver transplant data from OPTN, which was supported in part by Health Resources and Services Administration contract HHSH250-2019-00001C.

\bibliography{references}

\newpage
\appendix
\onecolumn

\section{Appendix A: Applicability of Gradient Descent}
\vspace{6pt}

The negative log-likelihood~$\lambda=-\log\mathcal{L}(\{r_{\theta_i}\};\mathcal{P})$ can be minimized using gradient descent as long as $r_{\theta}$ is a differentiable parameterization with respect to $\theta$ (i.e.\ when $\nabla_{\theta}r_{\theta}(x)$ is well-defined) since we have
\begin{align*}
    \nabla_{\theta_{\ell}} \lambda
    &= -\nabla_{\theta_{\ell}} \log\mathcal{L}(\{r_{\theta_i}\};\mathcal{P})\\
    &= -\nabla_{\theta_{\ell}} \frac{1}{2}\sum\nolimits_{(x^{\star},x^{\circ})\in X\times X}\bigg[ n(x^{\star},x^{\circ})\log\Pr(x^{\star}\succ x^{\circ}|\{r_{\theta_i}\})+n(x^{\circ},x^{\star})\log\Pr(x^{\circ}\succ x^{\star}|\{r_{\theta_i}\}) \bigg] \\
    &= -\nabla_{\theta_{\ell}} \sum\nolimits_{(x^{\star},x^{\circ})\in X\times X}\bigg[ n(x^{\star},x^{\circ})\log\Pr(x^{\star}\succ x^{\circ}|\{r_{\theta_i}\}) \bigg] \\
    &= -\sum\nolimits_{(x^{\star},x^{\circ})\in X\times X}\bigg[ n(x^{\star},x^{\circ}) \nabla_{\theta_{\ell}}\log\Pr(x^{\star}\succ x^{\circ}|\{r_{\theta_i}\}) \bigg] \\
    &= -\sum\nolimits_{(x^{\star},x^{\circ})\in X\times X}\bigg[ n(x^{\star},x^{\circ}) \frac{\nabla_{\theta_{\ell}}\Pr(x^{\star}\succ x^{\circ}|\{r_{\theta_i}\})}{\Pr(x^{\star}\succ x^{\circ}|\{r_{\theta_i}\})} \bigg] ~, \\[12pt]
    \nabla_{\theta_{\ell}}\Pr(x^{\star}\succ x^{\circ}|\{r_{\theta_i}\})
    &= \nabla_{\theta_{\ell}}\sum_{i=1}^k\bigg[ \Pr(x^{\star}\succ_i x^{\circ}|r_{\theta_i})\prod_{j=1}^{i-1}\Pr(x^{\star}\equiv_j x^{\circ}|r_{\theta_j}) \bigg] \\
    &= \bigg( \prod_{j=1}^{\ell-1}\Pr(x^{\star}\equiv_j x^{\circ}|r_{\theta_j})  \bigg)\cdot\nabla_{\theta_{\ell}}\Pr(x^{\star}\succ_{\ell} x^{\circ}|r_{\theta_{\ell}}) \\
    &\hspace{24pt}+ \bigg( \sum_{i=\ell+1}^{k}\Big[\Pr(x^{\star}\succ_i x^{\circ}|r_{\theta_i})\prod\nolimits_{j\in\{1,\ldots,i-1\},j\neq\ell}\Pr(x^{\star}\equiv_j x^{\circ}|r_{\theta_j})\Big] \bigg)\cdot\nabla_{\theta_{\ell}}\Pr(x^{\star}\equiv_{\ell} x^{\circ}|r_{\theta_{\ell}}) ~, \\[12pt]
    \nabla_{\theta_{\ell}}\Pr(x^{\star}\succ_{\ell} x^{\circ}|r_{\theta_{\ell}})
    &= \nabla_{\theta_{\ell}}\frac{1}{1+e^{-\alpha_{\ell}(r_{\theta_{\ell}}(x^{\star})-r_{\theta_{\ell}}(x^{\circ})-\varepsilon_{\ell})}} \\
    &= \bigg(\frac{1}{1+e^{-\alpha_{\ell}(r_{\theta_{\ell}}(x^{\star})-r_{\theta_{\ell}}(x^{\circ})-\varepsilon_{\ell})}}\bigg)\cdot\bigg(1-\frac{1}{1+e^{-\alpha_{\ell}(r_{\theta_{\ell}}(x^{\star})-r_{\theta_{\ell}}(x^{\circ})-\varepsilon_{\ell})}}\bigg) \\
    &\hspace{24pt}\times \alpha_{\ell}\big(\nabla_{\theta_{\ell}}r_{\theta_{\ell}}(x^{\star})-\nabla_{\theta_{\ell}}r_{\theta_{\ell}}(x^{\circ})\big) ~, \\[12pt]
    \nabla_{\theta_{\ell}}\Pr(x^{\star}\equiv_{\ell}x^{\circ}|r_{\theta_{\ell}}) &= 1 - \nabla_{\theta_{\ell}}\Pr(x^{\star}\succ_{\ell} x^{\circ}|r_{\theta_{\ell}}) - \nabla_{\theta_{\ell}}\Pr(x^{\circ}\succ_{\ell} x^{\star}|r_{\theta_{\ell}}) ~.
\end{align*}

 Moreover, the negative log-likelihood $\lambda$ is also differentiable with respect to parameters~$\{\alpha_i\}$ and $\{\varepsilon_i\}$---hence these parameters can be optimized by minimizing $\lambda$ as well. Similar to $\theta_{\ell}$, we have
 \begin{align*}
    \frac{\partial}{\partial\alpha_{\ell}}\lambda
    &= -\sum\nolimits_{(x^{\star},x^{\circ})\in X\times X}\bigg[ \frac{n(x^{\star},x^{\circ})}{\Pr(x^{\star}\succ x^{\circ}|\{r_{\theta_i}\})}\cdot\frac{\partial}{\partial\alpha_{\ell}}\Pr(x^{\star}\succ x^{\circ}|\{r_{\theta_i}\}) \bigg] ~, \\[12pt]
    \frac{\partial}{\partial\alpha_{\ell}}\Pr(x^{\star}\succ_{\ell} x^{\circ}|r_{\theta_{\ell}})
    &= \bigg( \prod_{j=1}^{\ell-1}\Pr(x^{\star}\equiv_j x^{\circ}|r_{\theta_j}) \bigg)\cdot\frac{\partial}{\partial\alpha_{\ell}}\Pr(x^{\star}\succ_{\ell} x^{\circ}|r_{\theta_{\ell}}) \\
    &\hspace{24pt}+ \bigg( \sum_{i=\ell+1}^{k}\Big[\Pr(x^{\star}\succ_i x^{\circ}|r_{\theta_i})\prod\nolimits_{j\in\{1,\ldots,i-1\},j\neq\ell}\Pr(x^{\star}\equiv_j x^{\circ}|r_{\theta_j})\Big] \bigg)\cdot\frac{\partial}{\partial\alpha_{\ell}}\Pr(x^{\star}\equiv_{\ell} x^{\circ}|r_{\theta_{\ell}}) ~, \\[12pt]
    \frac{\partial}{\partial\varepsilon_{\ell}}\lambda
    &= -\sum\nolimits_{(x^{\star},x^{\circ})\in X\times X}\bigg[ \frac{n(x^{\star},x^{\circ})}{\Pr(x^{\star}\succ x^{\circ}|\{r_{\theta_i}\})}\cdot\frac{\partial}{\partial\varepsilon_{\ell}}\Pr(x^{\star}\succ x^{\circ}|\{r_{\theta_i}\}) \bigg] ~, \\[12pt]
    \frac{\partial}{\partial\varepsilon_{\ell}}\Pr(x^{\star}\succ_{\ell} x^{\circ}|r_{\theta_{\ell}})
    &= \bigg( \prod_{j=1}^{\ell-1}\Pr(x^{\star}\equiv_j x^{\circ}|r_{\theta_j}) \bigg)\cdot\frac{\partial}{\partial\varepsilon_{\ell}}\Pr(x^{\star}\succ_{\ell} x^{\circ}|r_{\theta_{\ell}}) \\
    &\hspace{24pt}+ \bigg( \sum_{i=\ell+1}^{k}\Big[\Pr(x^{\star}\succ_i x^{\circ}|r_{\theta_i})\prod\nolimits_{j\in\{1,\ldots,i-1\},j\neq\ell}\Pr(x^{\star}\equiv_j x^{\circ}|r_{\theta_j})\Big] \bigg)\cdot\frac{\partial}{\partial\varepsilon_{\ell}}\Pr(x^{\star}\equiv_{\ell} x^{\circ}|r_{\theta_{\ell}}) ~.
\end{align*}
Then,
\begin{align*}
    \frac{\partial}{\partial\alpha_{\ell}}\Pr(x^{\star}\succ_{\ell} x^{\circ}|r_{\theta_{\ell}})
    &= \frac{\partial}{\partial\alpha_{\ell}}\bigg( \frac{1}{1+e^{-\alpha_{\ell}(r_{\theta_{\ell}}(x^{\star})-r_{\theta_{\ell}}(x^{\circ})-\varepsilon_{\ell})}} \bigg) \\
    &= \bigg(\frac{1}{1+e^{-\alpha_{\ell}(r_{\theta_{\ell}}(x^{\star})-r_{\theta_{\ell}}(x^{\circ})-\varepsilon_{\ell})}}\bigg)\cdot\bigg(1-\frac{1}{1+e^{-\alpha_{\ell}(r_{\theta_{\ell}}(x^{\star})-r_{\theta_{\ell}}(x^{\circ})-\varepsilon_{\ell})}}\bigg) \\
    &\hspace{24pt}\times (r_{\theta_{\ell}}(x^{\star})-r_{\theta_{\ell}}(x^{\circ})-\varepsilon_{\ell}) ~, \\[12pt]
    \frac{\partial}{\partial\alpha_{\ell}}\Pr(x^{\star}\equiv_{\ell}x^{\circ}|r_{\theta_{\ell}}) &= 1 - \frac{\partial}{\partial\alpha_{\ell}}\Pr(x^{\star}\succ_{\ell} x^{\circ}|r_{\theta_{\ell}}) - \frac{\partial}{\partial\alpha_{\ell}}\Pr(x^{\circ}\succ_{\ell} x^{\star}|r_{\theta_{\ell}}) ~, \\[12pt]
    \frac{\partial}{\partial\varepsilon_{\ell}}\Pr(x^{\star}\succ_{\ell} x^{\circ}|r_{\theta_{\ell}})
    &= \frac{\partial}{\partial\varepsilon_{\ell}}\bigg( \frac{1}{1+e^{-\alpha_{\ell}(r_{\theta_{\ell}}(x^{\star})-r_{\theta_{\ell}}(x^{\circ})-\varepsilon_{\ell})}} \bigg) \\
    &= -\alpha_{\ell}\cdot\bigg(\frac{1}{1+e^{-\alpha_{\ell}(r_{\theta_{\ell}}(x^{\star})-r_{\theta_{\ell}}(x^{\circ})-\varepsilon_{\ell})}}\bigg)\cdot\bigg(1-\frac{1}{1+e^{-\alpha_{\ell}(r_{\theta_{\ell}}(x^{\star})-r_{\theta_{\ell}}(x^{\circ})-\varepsilon_{\ell})}}\bigg) ~, \\[12pt]
    \frac{\partial}{\partial\varepsilon_{\ell}}\Pr(x^{\star}\equiv_{\ell}x^{\circ}|r_{\theta_{\ell}}) &= 1 - \frac{\partial}{\partial\varepsilon_{\ell}}\Pr(x^{\star}\succ_{\ell} x^{\circ}|r_{\theta_{\ell}}) - \frac{\partial}{\partial\varepsilon_{\ell}}\Pr(x^{\circ}\succ_{\ell} x^{\star}|r_{\theta_{\ell}}) ~.
\end{align*}

Note that computing the gradient with respect to the parameters of the $i$-th reward function requires computing the gradients of $i$-many sigmoid functions. Hence, the complexity of LORI should scale with $O(k^2)$.

\vspace{6pt}
\section{Appendix B: Additional Experiments}
\vspace{6pt}

\subsection{Additional Experiments for Improving Behavior in Cancer Treatment}

We repeat the experiments for the cancer treatment setting, but instead of $r_1$, $r_2$ in \eqref{eqn:cancer-rewards}, the ground-truth preferences are induced by the single-dimensional reward function
\begin{equation}
        r(\bm{a}_{1:\tau},\bm{z}_{1:\tau},\bm{w}_{1:\tau}) = -0.2\times\frac{1}{\tau}\sum_{t=1}^{\tau}z_t + 0.8\times\frac{1}{\tau}\sum_{t=1}^{\tau}w_t
\end{equation}
according to the single-dimensional preference model in \eqref{eqn:model-single} with $\alpha=10\log 9$. Similar to the original setting, WBC still has higher importance (with weight $0.8$) than the tumor volume (with weight $-0.2$) but keeping it above a threshold of five is no longer lexicographically prioritized.

We consider the same benchmarks as before. In particular, we still run LORI with $k=2$ and infer two lexicographically-ordered reward functions to represent preferences despite the ground-truth reward function being single-dimensional. The results are reported in Tables~\ref{tbl:reward-single} and \ref{tbl:policy-single}. We see that LORI still performs comparably to T-REX in terms of accuracy in predicting preferences (with no statistically significant difference between the two methods) and performs the best in terms of frequency of improvements over the behavioral policy. This is expected as, in LORI, preference models with $k=2$ are strictly a generalization of preference models with $k=1$; setting $\varepsilon_1=0$ in the former models recovers the latter models.

\begin{table*}[h]
    \centering
    \caption{Comparison of reward functions based on preference prediction performance (in the cancer treatment setting with the single-dimensional ground-truth reward function).}
    \label{tbl:reward-single}
    \small
    \begin{tabular}{@{}l*2{c}@{}}
        \toprule
        \bf Algorithm & \bf \makecell{RMSE} & \bf \makecell{Accuracy} \\
        \midrule
        BIRL & $0.142\!\pm\! 0.038$ & $94.5\%\!\pm\! 0.24\%$ \\
        T-REX & $0.015\!\pm\! 0.004$ & $97.0\%\!\pm\! 0.21\%$ \\
        LORI & $0.025\!\pm\! 0.011$ & $96.9\%\!\pm\! 0.32\%$ \\
        \bottomrule
    \end{tabular}
\end{table*}
\begin{table*}[h]
    \centering
    \caption{Comparison of policies based on how often the row policies are preferred to the column policies (in the cancer treatment setting with the single-dimensional ground-truth reward function).}
    \label{tbl:policy-single}
    \small
    \begin{tabular}{@{}l|*5{c}@{}}
        \toprule
            & \bf Behavior & \bf BC & \bf BIRL & \bf T-REX & \bf LORI \\
        \midrule
        BC & $48.8\%\!\pm\! 2.17\%$ & -- \\
        BIRL & $68.2\%\!\pm\! 1.39\%$ & $69.6\%\!\pm\! 0.96\%$ & -- \\
        T-REX & $72.8\%\!\pm\! 1.36\%$ & $73.7\%\!\pm\! 1.00\%$ & $55.3\%\!\pm\! 1.26\%$ & -- \\
        LORI & $79.0\%\!\pm\! 1.29\%$ & $78.9\%\!\pm\! 0.43\%$ & $61.5\%\!\pm\! 1.06\%$ & $56.3\%\!\pm\! 1.75\%$ & -- \\
        Optimal & $74.4\%\!\pm\! 1.70\%$ & $74.7\%\!\pm\! 1.32\%$ & $57.2\%\!\pm\! 0.74\%$ & $52.4\%\!\pm\! 1.07\%$ & $45.9\%\!\pm\! 1.25\%$ \\
        \bottomrule
    \end{tabular}
\end{table*}

\subsection{Additional Experiments Regarding the Number of Reward Functions}

In the main paper, we mentioned how using lexicographic models that only employ a small number of reward functions would be enough in most cases. Because in lexicographic models, even when the number objectives that is considered by the agent is large, the number of objectives that actually affect the outcome for a particular decision tends to be small. For instance, consider an agent with $k=10$ reward functions. While the first reward functions is relevant for almost all decisions, the last reward functions becomes relevant only when the alternatives are deemed equal in terms of the all first nine reward functions; this happens only with probability~$\prod_{i=1}^9 \Pr(x^{\star}\equiv_i x^{\circ})$. In general, the less important a reward function is, less likely it becomes for it to have a significant effect on preference. Hence, accuracy gained by adding more reward functions to a model gets smaller and smaller as more reward functions are added.

We demonstrate this empirically in a synthetic environment. Let the set of alternatives be $X=\mathbb{R}^{10}$ and suppose an agent makes preferences based on $k^{\mathrm{true}}=10$ reward functions. Let these reward functions be given in a linear form such that $r_i(x)=\theta_i \cdot x$ where $\theta_i\in \mathbb{R}_{+}^{10}$, and $\|\theta_i\|_1=1$ for all $i\in\{1,\ldots,k^{\mathrm{true}}\}$. We sample $\theta_i$'s uniformly at random, sample $\varepsilon_i$'s from the standard half-normal distribution, and set $\alpha_i=5\log 4$ for all $i\in\{1,\ldots,k^{\mathrm{true}}\}$. We generate preference data~$\mathcal{P}$ by sampling and evaluating $10,\!000$ pairs of alternatives, where each alternative~$x$ is sampled from the multivariate Gaussian distribution~$\mathcal{N}(\bm{0},0.5^2I)$. Then, we infer representations of the agent's preferences using LORI with varying number of reward functions $k=\{1,2,\ldots,10\}$ while the true number of reward functions is always $k^{\mathrm{true}}=10$. We repeat this experiment five times to obtain error bars.

\begin{figure}[h]
    \centering
    \includegraphics[width=.33\linewidth]{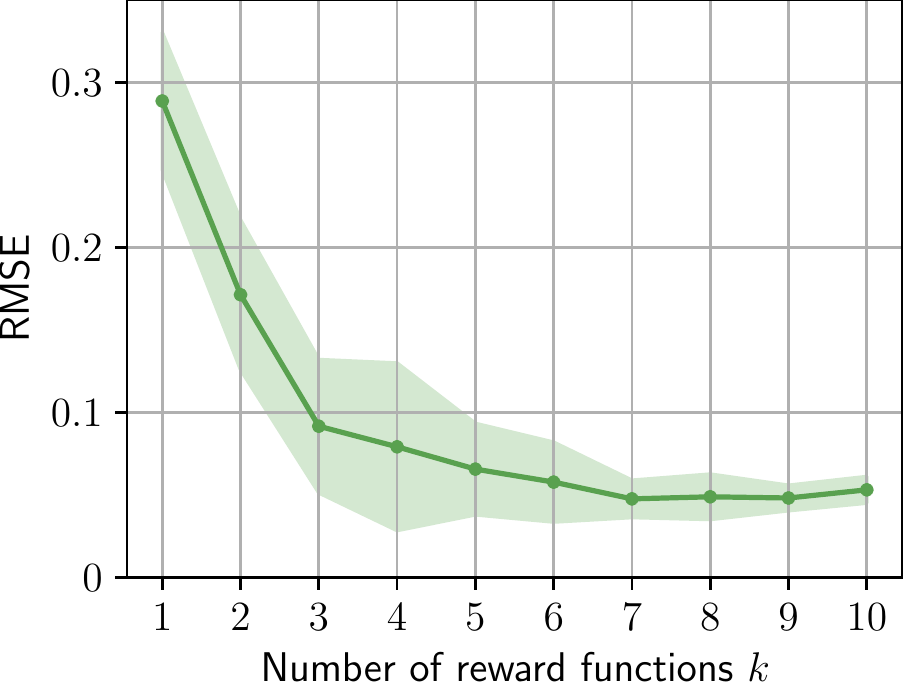}
    \caption{RMSE of representations with different number of reward functions~$k$ (note that $k^{\mathrm{true}}=10$).}
    \label{fig:number}
\end{figure}

We measure the RMSE of the preference probabilities estimated by the inferred representations using a test set of $10,\!000$ additional preferences. Figure~\ref{fig:number} plots the RMSE of each representation with respect to the number of reward functions~$k$ they employ. As expected, the RMSE drops significantly when $k$ is increased from one to two and two to three. As we add more reward functions, the RMSE keeps dropping but less dramatically until we reach $k=7$. When we use more than $k=7$ reward functions, the RMSE does not improve even though the true number of reward functions is $k^{\mathrm{true}}=10$. This is because the fine adjustments made by reward functions $8$, $9$, and $10$ become smaller than the noise in the data.

\subsection{Additional Experiments Where Factors Determining Priorities Are Identified}

There are settings in which the priorities given to objectives depend on specifics of the situation. In this section, we give one such example and show how LORI can be used to identify under what circumstances one objective is prioritized over another. Consider the choice of treatment plan for a life-threatening medical condition. Suppose that various protocols vary according to their efficacy and toxicity (side-effects), and that---as would seem natural---efficacy is prioritized for younger patients and toxicity is priorities for older patients.

We simulate cancer treatment plans as before, and consider tumor volume as a measure of efficacy and WBC count as a measure of toxicity. Different from before, we sample the age of the patient who would hypothetically receive the treatment from $\mathcal{N}(35,20^2)$ (we denote it with $y$). We set the reward functions $r_1$, $r_2$ such that clinicians prioritize maximizing efficacy for patients under $y_{\text{threshold}}=40$ years old:
\begin{align*}
    r_1(\bm{a}_{1:\tau},y,\bm{z}_{1:\tau},\bm{w}_{1:\tau}) &= \frac{1}{1+e^{-(y-y_{\text{threshold}})/y_{\text{sensitivity}}}} \cdot \bigg(-\frac{1}{\tau}\sum_{t=1}^{\tau}z_t\bigg) + \bigg(1-\frac{1}{1+e^{-(y-y_{\text{threshold}})/y_{\text{sensitivity}}}}\bigg) \cdot \frac{1}{\tau}\sum_{t=1}^{\tau}w_t ~, \\
    r_2(\bm{a}_{1:\tau},y,\bm{z}_{1:\tau},\bm{w}_{1:\tau}) &= \bigg(1-\frac{1}{1+e^{-(y-y_{\text{threshold}})/y_{\text{sensitivity}}}}\bigg) \cdot \bigg(-\frac{1}{\tau}\sum_{t=1}^{\tau}z_t\bigg) + \frac{1}{1+e^{-(y-y_{\text{threshold}})/y_{\text{sensitivity}}}} \cdot \frac{1}{\tau}\sum_{t=1}^{\tau}w_t ~,
\end{align*}
where $y_{\text{sensitivity}}=1$ determines the clinicians' sensitivity to age. Here, the ratio $1/(1+e^{-(y-y_{\text{threshold}})/y_{\text{sensitivity}}}) \in (0,1)$ can be thought of as a measure of how much efficacy is prioritized over toxicity.

\begin{figure}[h]
    \centering
    \includegraphics[width=.33\linewidth]{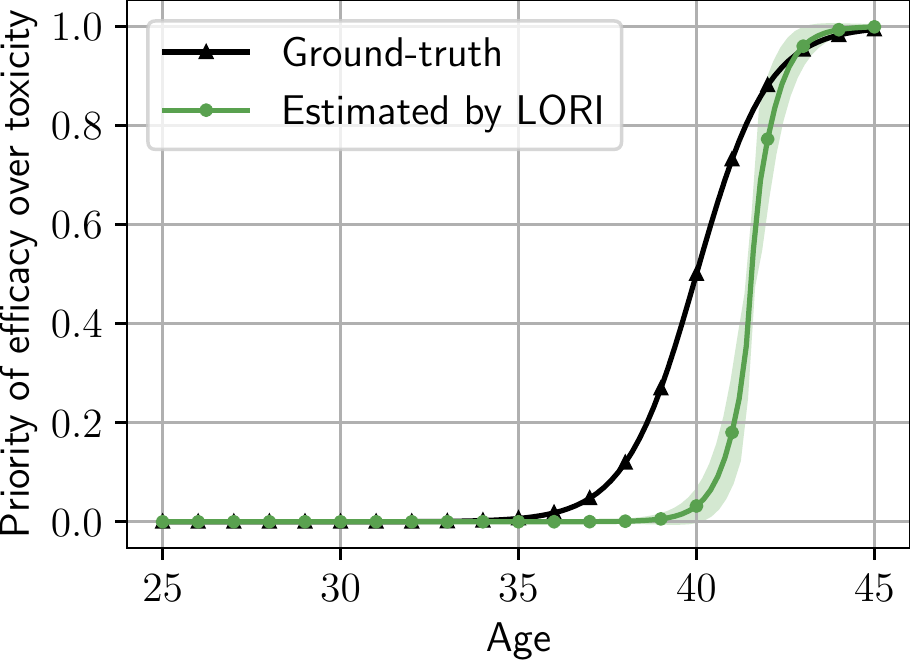}
    \caption{Priority of efficacy over toxicity with respect to a patient's age.}
    \label{fig:age}
\end{figure}

We generate 1000 trajectories with $\tau=20$ and generate preferences according to $r_1$, $r_2$ by sampling 1000 pairs from the generated trajectories. Given the generated preferences, we use LORI to identify $y_{\text{threshold}}$ and $y_{\text{sensitivity}}$. Figure~\ref{fig:age} shows the estimated priority of efficacy over toxicity---as measured by $1/(1+e^{-(y-y_{\text{threshold}})/y_{\text{sensitivity}}})$---with respect to age as well as the ground-truth priority of efficacy over toxicity.

\vspace{6pt}
\section{Appendix C: Experimental Details}
\vspace{6pt}

\subsection{Details of improving behavior in cancer}

\paragraph{BC}
We train a neural network whose inputs are the patient features~$z_t$ and $w_t$ at time~$t$, and whose output is the predicted probabilities~$\hat{\pi}_b(a|z_t,w_t)$ of taking action~$a$ given the input features. Given dataset~$\mathcal{D}$, we minimize the cross-entropy loss~$\mathcal{L}=-\sum_{\bm{a}_{1:\tau},\bm{z}_{1:\tau},\bm{w}_{1:\tau}\in\mathcal{D}}\sum_{t=1}^{\tau}\mathbb{I}\{a_t=a\}\log\hat{\pi}_b(a|z_t,w_t)$ using RMSprop with learning rate $0.001$ and discount rate $0.9$ until convergence---that is when the cross-entropy loss does not improve for $100$ consecutive iterations.

\paragraph{BIRL}
We assume that the true reward function has the linear form 
\begin{align*}
    r(\bm{a}_{1:\tau},\bm{z}_{1:\tau},\bm{w}_{1:\tau})=\frac{1}{\tau}\sum_{t=1}^{\tau}\left(-\theta^{(z)}z_t+\theta^{(w)}w_t\right) ~,
\end{align*}
where $\theta^{(z)}\in\mathbb{R}_+$ and $\theta^{(w)}\in\mathbb{R}_+$ are unknown parameters to be estimated. We use the Markov chain Monte Carlo (MCMC) algorithm proposed by \citet{ramachandran2007bayesian}. Given the last samples~$\theta^{(z)}$, $\theta^{(w)}$; new candidate samples $\theta^{(z)}{}',\theta^{(w)}{}'$ are generated as $\theta^{(z)}{}'=\theta^{(z)}e^{\eta^{(z)}}$ and $\theta^{(w)}{}'=\theta^{(w)}e^{\eta^{(w)}}$, where $\eta^{(z)},\eta^{(w)}\sim\mathcal{N}(0,0.01^2)$. A final estimate is formed by averaging every $100$th sample among $10,\!000$ total samples (ignoring the initial $1000$ samples). 

\paragraph{T-REX}
We assume that the true reward function has the same linear form as in BIRL. Then, as \citet{brown2019extrapolating} proposes, we minimize the negative log-likelihood $-\log\mathcal{L}$ defined in \eqref{eqn:likelihood-single} using RMSprop with learning rate $0.001$ and discount rate $0.9$ until convergence---that is when the negative log-likelihood does not improve for $10$ consecutive iterations.

\paragraph{LORI}
We assume that the true reward functions~$r_i\in\{r_1,r_2\}$ have the thresholded linear form
\begin{align*}
    r_i(\bm{a}_{1:\tau},\bm{z}_{1:\tau},\bm{w}_{1:\tau})=\softmin\left\{\theta_i^{(\max)}, \frac{1}{\tau}\sum_{t=1}^{\tau}\left(-\theta_i^{(z)}z_t+\theta_i^{(w)}w_t\right)\right\} ~,
\end{align*}
where $\theta_i^{(\max)}\in\mathbb{R}$, $\theta_i^{(z)}\in\mathbb{R}_+$, and $\theta_i^{(w)}\in\mathbb{R}_+$ are unknown parameters to be estimated together with $\varepsilon_i\in\mathbb{R}_{+}$. We minimize the negative log-likelihood~$\lambda=-\log\mathcal{L}$ using RMSprop with learning rate $0.001$ and discount rate $0.9$ until convergence---that is when the negative log-likelihood does not improve for $10$ consecutive iterations. Algorithm~\ref{alg:cancer} outlines the complete learning procedure for LORI.

\begin{algorithm}
    \caption{LORI for improving behavior in cancer treatment}
    \label{alg:cancer}
    \begin{algorithmic}[1]
        \STATE \textbf{Input:} Preferences~$\mathcal{P}$
        \STATE \textbf{Parameters:} Learning rate $\eta=0.001$, discount rate $\gamma=0.9$
        \STATE $\theta_1^{(\max)}, \log\theta_1^{(z)}, \log\theta_1^{(w)}, \log\varepsilon_1, \theta_2^{(\max)}, \log\theta_2^{(z)}, \log\theta_2^{(w)}, \log\varepsilon_2\gets 0$
        \STATE $H_{\theta_1^{(\max)}}, H_{\log\theta_1^{(z)}}, H_{\log\theta_1^{(w)}}, H_{\log\varepsilon_1}, H_{\theta_2^{(\max)}}, H_{\log\theta_2^{(z)}}, H_{\log\theta_2^{(w)}}, H_{\log\varepsilon_2}\gets 0$
        \FOR{$t\in\{1,2,\ldots\}$}
            \FOR{$\phi\in\{\theta_j^{(\max)},\log\theta_j^{(z)},\log\theta_j^{(w)},\log\varepsilon_j\}_{j\in\{1,2\}}$}
                \STATE $G_{\phi}\gets \nabla_{\phi}\lambda(\{\theta_j^{(\max)},\theta_j^{(z)},\theta_j^{(w)},\varepsilon_j\}_{j\in\{1,2\}})$
                \STATE $H_{\phi}\gets \gamma H_{\phi} + (1-\gamma)G_{\phi}^2$
            \ENDFOR
            \FOR{$\phi\in\{\theta_j^{(\max)},\log\theta_j^{(z)},\log\theta_j^{(w)},\log\varepsilon_j\}_{j\in\{1,2\}}$}
                \STATE $\phi\gets\phi -\eta\,G_{\phi}/\sqrt{H_{\phi}}$
            \ENDFOR
            \STATE $\lambda^{(t)}\gets \lambda(\{\theta_j^{(\max)},\theta_j^{(z)},\theta_j^{(w)},\varepsilon_j\}_{j\in\{1,2\}})$
            \STATE \textbf{if} $\lambda^{(t)}>\lambda^{(t-10)}$ \textbf{then break}
        \ENDFOR
        \STATE \textbf{Output:} Parameters~$\{\theta_j^{(\max)},\theta_j^{(z)},\theta_j^{(w)},\varepsilon_j\}_{j\in\{1,2\}}$
    \end{algorithmic}
\end{algorithm}

\subsection{Details of understanding behavior in organ transplantation}

\paragraph{Dataset}
In order to estimate the number of days patients would survive without a transplant, we train a Cox model (M1) using the following features: Age, Gender, Creatinine, Bilirubin, INR, Sodium, and Bilirubin~$\times$~Sodium. In order to estimate the number of days a patient would survive after a transplantation, we train another Cox model (M2) using the following features: Age, Gender, HCV, Creatinine, Bilirubin, INR, Sodium, Albumin, Previous abdominal surgery, Encephalopathy, Ascites, Waiting time, Diabetes, Donor age, Donor cause of death, Donor BMI, Donor diabetes, ABO compatibility, HCV~$\times$~Donor diabetes, HCV~$\times$~Donor age, Age~$\times$~Creatinine. These are the same features used by TransplantBenefit except those that were not available in the OPTN data. When training both models, we only consider patients who were in the waitlist during 2019 and filter out patients with missing features (similar to our main analysis). Then, benefit is defined as the difference between the estimates of M2 and M1 and need is defined as the negative of the estimate of M1.

\paragraph{T-REX \& LORI}
As before, we infer reward functions (and $\varepsilon_i$'s in the case of LORI) by minimizing the negative log-likelihood using RMSprop with learning rate $0.001$ and discount rate $0.9$ until convergence---that is when the log-likelihood does not improve for $10$ consecutive iterations. Letting $r_i=\theta_i^{(B)}\cdot\textit{Benefit}+\theta_i^{(N)}\cdot\textit{Need}$, Algorithm~\ref{alg:organ} outlines the complete learning procedure for LORI. All experiments are run on a personal computer with an Intel Core i9 processor.

\begin{algorithm}
    \caption{LORI for understanding behavior in organ transplantation}
    \label{alg:organ}
    \begin{algorithmic}[1]
        \STATE \textbf{Input:} Preferences~$\mathcal{P}$
        \STATE \textbf{Parameters:} Learning rate $\eta=0.001$, discount rate $\gamma=0.9$
        \STATE $\log\theta_1^{(B)}, \log\theta_1^{(N)}, \log\varepsilon_1, \log\theta_2^{(B)}, \log\theta_2^{(N)}, \log\varepsilon_2\gets 0$
        \STATE $H_{\log\theta_1^{(B)}}, H_{\log\theta_1^{(N)}}, H_{\log\varepsilon_1}, H_{\log\theta_2^{(B)}}, H_{\log\theta_2^{(N)}}, H_{\log\varepsilon_2}\gets 0$
        \FOR{$t\in\{1,2,\ldots\}$}
            \FOR{$\phi\in\{\log\theta_j^{(B)},\log\theta_j^{(N)},\log\varepsilon_j\}_{j\in\{1,2\}}$}
                \STATE $G_{\phi}\gets \nabla_{\phi}\lambda(\{\theta_j^{(B)},\theta_j^{(N)},\varepsilon_j\}_{j\in\{1,2\}})$
                \STATE $H_{\phi}\gets \gamma H_{\phi} + (1-\gamma)G_{\phi}^2$
            \ENDFOR
            \FOR{$\phi\in\{\log\theta_j^{(B)},\log\theta_j^{(N)},\log\varepsilon_j\}_{j\in\{1,2\}}$}
                \STATE $\phi\gets\phi -\eta\,G_{\phi}/\sqrt{H_{\phi}}$
            \ENDFOR
            \STATE $\lambda^{(t)}\gets \lambda(\{\theta_j^{(B)},\theta_j^{(N)},\varepsilon_j\}_{j\in\{1,2\}})$
            \STATE \textbf{if} $\lambda^{(t)}>\lambda^{(t-10)}$ \textbf{then break}
        \ENDFOR
        \STATE \textbf{Output:} Parameters~$\{\theta_j^{(B)},\theta_j^{(N)},\varepsilon_j\}_{j\in\{1,2\}}$
    \end{algorithmic}
\end{algorithm}

\end{document}